\documentclass[final]{cvpr}

\usepackage{times}
\usepackage{enumitem}
\usepackage{epsfig}
\usepackage{graphicx}
\usepackage{amsmath}
\usepackage{amssymb}
\usepackage{caption}
\usepackage{booktabs}
\usepackage{array,multirow}
\usepackage{soul}
\usepackage{pifont}
\usepackage{slashbox}
\makeatletter
\@namedef{ver@eso-pic.sty}{}
\makeatother
\usepackage{pdfpages}

\usepackage[numbers,sort]{natbib} 

\newcommand{\cmark}{\ding{51}}%
\usepackage[pagebackref=true,breaklinks=true,colorlinks,bookmarks=false]{hyperref}

\def\cvprPaperID{7531} 
\def\confYear{CVPR 2021}
\begin{document}

\title{HVPR: Hybrid Voxel-Point Representation for Single-stage 3D Object Detection}

\author{Jongyoun Noh \quad\quad\quad Sanghoon Lee \quad\quad\quad Bumsub Ham\thanks{Corresponding author.}\vspace*{0.2cm}\\
{School of Electrical and Electronic Engineering, Yonsei University}}

\maketitle
\thispagestyle{empty}

\begin{abstract}
We address the problem of 3D object detection, that is, estimating 3D object bounding boxes from point clouds. 3D object detection methods exploit either voxel-based or point-based features to represent 3D objects in a scene. Voxel-based features are efficient to extract, while they fail to preserve fine-grained 3D structures of objects. Point-based features, on the other hand, represent the 3D structures more accurately, but extracting these features is computationally expensive. We introduce in this paper a novel single-stage 3D detection method having the merit of both voxel-based and point-based features. To this end, we propose a new convolutional neural network~(CNN) architecture, dubbed HVPR, that integrates both features into a single 3D representation effectively and efficiently. Specifically, we augment the point-based features with a memory module to reduce the computational cost. We then aggregate the features in the memory, semantically similar to each voxel-based one, to obtain a hybrid 3D representation in a form of a pseudo image, allowing to localize 3D objects in a single stage efficiently. We also propose an Attentive Multi-scale Feature Module~(AMFM) that extracts scale-aware features considering the sparse and irregular patterns of point clouds. Experimental results on the KITTI dataset demonstrate the effectiveness and efficiency of our approach, achieving a better compromise in terms of speed and accuracy. 

\end{abstract}
\begin{figure}
\captionsetup{font={small}}
\centering
   \includegraphics[width=1.0\linewidth]{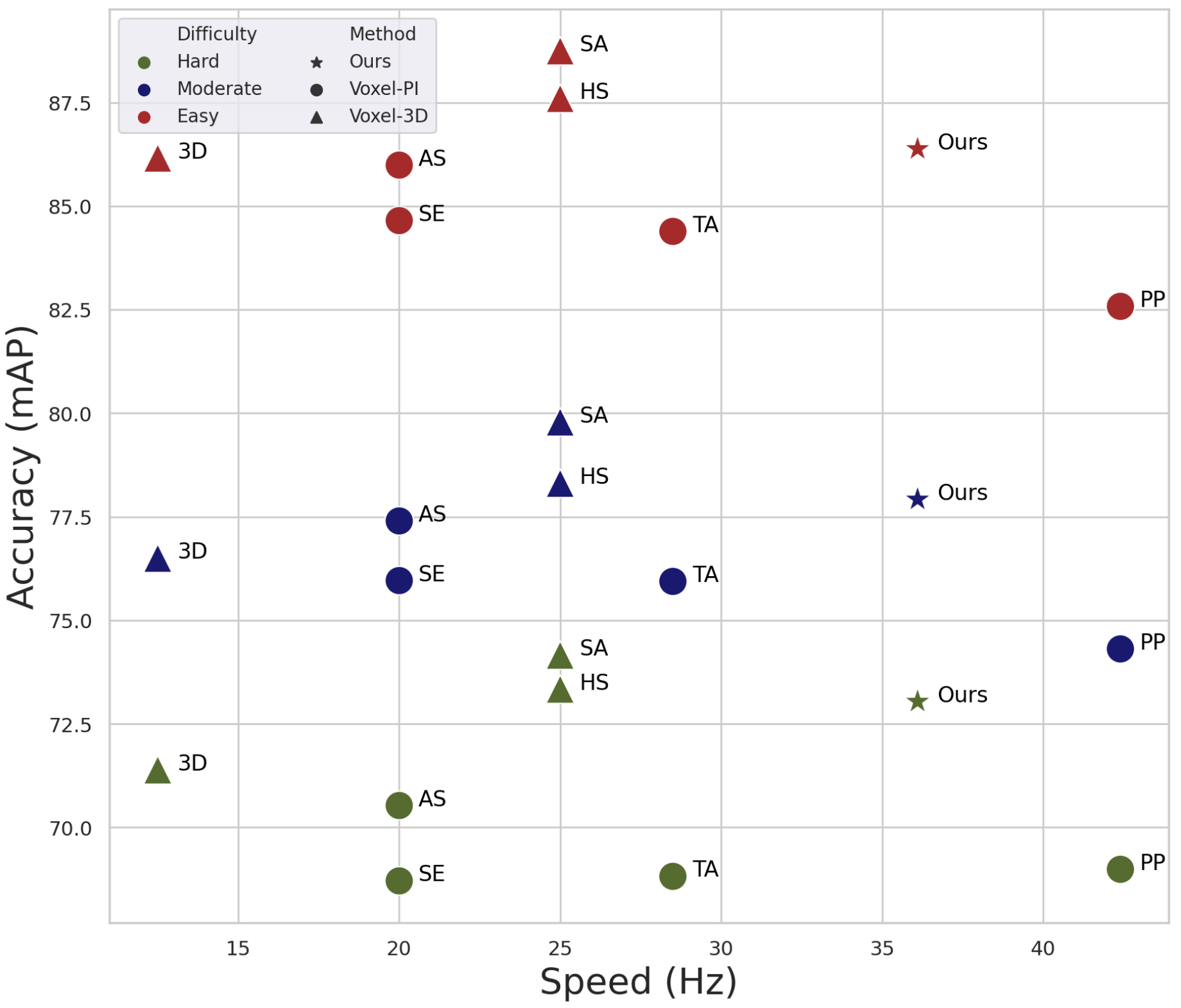}
   \vspace{-0.5cm}
   \caption{Runtime and accuracy comparison of detection results on the KITTI~\cite{geiger2012we} test set. We compare our model with voxel-based methods on the car class for three difficulty levels. Voxel-based methods using pseudo image representations (Voxel-PI) are shown as circles, and 3D voxel-based methods~(Voxel-3D) are plotted as triangles. Our method gives a better compromise in terms of accuracy and runtime for all cases. SE: SECOND~\cite{yan2018second}; PP: PointPillars~\cite{lang2019pointpillars}; TA: TANet~\cite{liu2020tanet}; AS: Associate-3D~\cite{du2020associate}; 3D: 3DIoULoss~\cite{zhou2019iou}; SA: SA-SSD~\cite{he2020structure}; HS: HotSpotNet~\cite{chen2019object}. Best viewed in color.}   

\vspace{-0.6cm}
\end{figure}

\section{Introduction}
3D object detection is an essential technique for scene understanding, which aims at predicting 3D bounding boxes of objects in a scene. It can also be exploited as a basic building block for many applications, including autonomous driving and robotics. Recent approaches to 3D object detection focus on learning discriminative 3D representations using point clouds acquired from a LiDAR sensor. Point clouds provide accurate depth information of objects, but they are sparse with the densities largely varying w.r.t distances from the sensor.

There are many attempts to learn 3D feature representations using point clouds with deep neural networks, which can be categorized into two groups: Voxel-based and point-based methods. The first approaches~\cite{du2020associate,he2020structure,lang2019pointpillars,yang2018pixor,ye2020hvnet,simony2018complex} transform raw point clouds to structured grid representations,~\eg,~birds-eye-view~(BEV)~\cite{yang2018pixor,simony2018complex} or voxels~\cite{he2020structure,lang2019pointpillars,shi2020pv,ye2020hvnet,zhou2018voxelnet}, to extract 3D representations using convolutional neural networks~(CNNs). The voxelization and downsampling operations in these methods allow to extract compact 3D features efficiently, but they fail to preserve fine-grained 3D structures of objects. The second approaches~\cite{qi2018frustum,shi2019pointrcnn,wang2019frustum,yang2019std,yang20203dssd}, on the other hand, exploit raw point clouds directly to extract point-wise features using~\eg,~PointNet++~\cite{qi2017pointnet,qi2017pointnet++}. They provide more discriminative 3D representations than the voxel-based one, and thus give better detection results. Processing a large-scale point cloud data, however, needs lots of computational cost.

We present in this paper a novel single-stage 3D object detection framework that integrates voxel-based and point-based features effectively and efficiently to obtain discriminative 3D representations. To this end, we introduce a new CNN architecture, dubbed HVPR, that consists of a two-stream encoder for voxel-based and point-based features and a memory module. It augments the encoder for point-based features with the memory module to reduce the computational cost. Namely, we update and store point-based features from the encoder to memory items during training, and do not use the encoder at test time, avoiding heavy computation. Specifically, we aggregate the point-wise features in memory items, semantically similar to each voxel-based one, to obtain a pseudo image representation, enabling exploiting hybrid 3D representations efficiently for localizing objects in a single stage. This also encourages voxel-based features to incorporate fine-grained representations of point-based features, which are particularly effective for localizing small or largely occluded objects acquired from sparse point clouds. We also introduce a detection network with an Attentive Multi-scale Feature Module~(AMFM). Given the hybrid pseudo image, it extracts multi-scale features, and AMFM refines them using 3D scale representations to obtain scale-aware features, which are crucial especially for 3D object detection, due to the sparse and irregular patterns of point clouds. Extensive experimental results on standard benchmarks demonstrate the effectiveness and efficiency of our approach to exploiting hybrid feature representations. The main contributions of this paper can be summarized as follows:
\vspace{-0.2cm}
\begin{itemize}[leftmargin=*]
   \item[$\bullet$] We introduce a novel single-stage framework for 3D object detection using hybrid 3D representations. We propose to use a memory module to augment point-based features, maintaining the efficiency of a single-stage method. 
   \vspace{-0.2cm}
   \item[$\bullet$] We introduce AMFM that provides scale-aware features considering the sparse and irregular patterns of point clouds explicitly, allowing to consider complex scale variations across objects for 3D object detection. \vspace{-0.2cm}
   \item[$\bullet$] We demonstrate that our approach gives a better compromise in terms of speed and accuracy, compared to the state of the art. Our model runs at 36.1fps, while achieving competitive performance on the KITTI dataset~\cite{geiger2012we}. \vspace{-0.2cm}
\end{itemize}
Our code and models are available online:~\url{https://cvlab.yonsei.ac.kr/projects/HVPR}.

\section{Related work}
\vspace{-0.2cm}

   \paragraph{Multi-sensor based 3D detection.}
      Multi-sensory data, obtained from~\eg,~RGB and depth sensors, provides complementary information for 3D object detection. For example, RGB and depth images give semantic and structural information of objects, respectively. Many approaches attempt to leverage features from RGB images and point clouds jointly for 3D object detection. MV3D~\cite{chen2017multi} generates object proposals from BEV representations, and refines them using features from point clouds and RGB images. AVOD~\cite{ku2018joint} instead incorporates these features to extract object proposals, which provides better detection results. To further enhance 3D representations, proxy tasks are also exploited~\cite{liang2019multi}, such as a ground estimation with point clouds and a depth completion with RGB images. Recent methods~\cite{huang2020epnet, yoo20203d} also handle the case when multi-sensory data is not registered. They leverage correspondences between features from point clouds and RGB images to better exploit the complementary information. 
      \vspace{-0.4cm}

\begin{figure*}[t]
   \captionsetup{font={small}}
      \centering
         \includegraphics[width=1.0\linewidth]{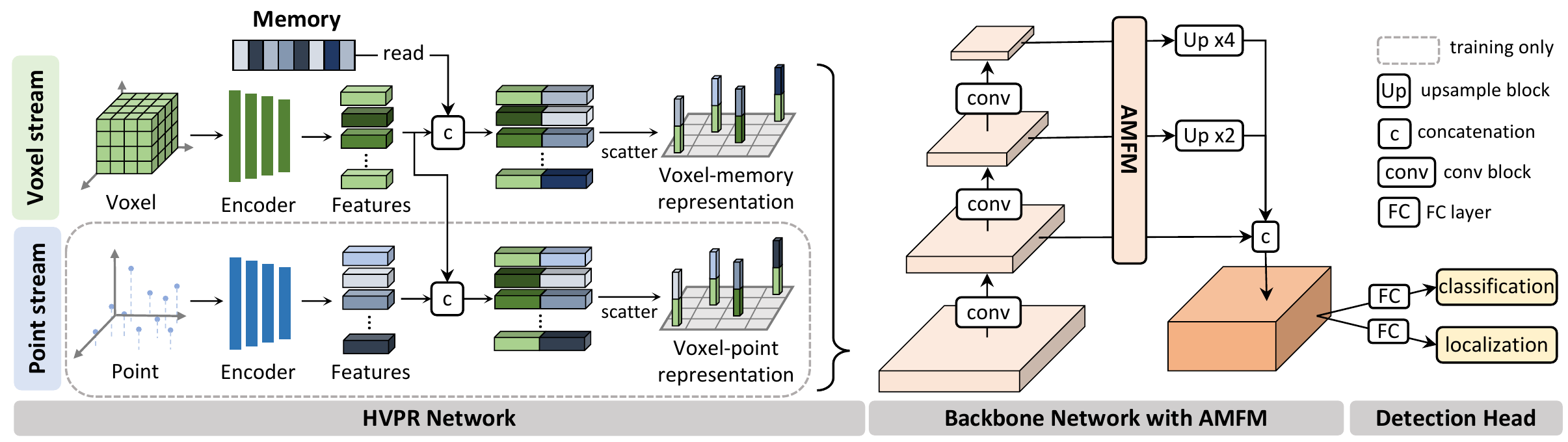}
\vspace{-0.5cm}
 \caption{An overview of our framework. The HVPR network inputs point clouds and generates two types of hybrid 3D features via a two-stream encoder: Voxel-point and voxel-memory representations. The former representations are obtained by aggregating point-based features for individual voxel-based ones. For the later ones, we also perform the aggregation but with memory items, instead of using point-based features. That is, we augment the point-based features using a memory module, and exploit voxel-memory representations,~\ie,~hybrid 3D features, at test time for fast inference. The backbone network with AMFM inputs the voxel-memory representations to extract multiple scale-aware features, and the detection head predicts 3D bounding boxes and object classes. See text for details.}

      \vspace{-0.45cm}
      \label{fig:overview}
    \end{figure*}    
\paragraph{LiDAR based 3D detection.}
      Recent approaches to 3D object detection use point clouds alone from a LiDAR sensor. They can be divided into two categories, voxel-based and point-based methods, depending on how to transform point clouds to 3D representations for localizing objects. First, voxel-based approaches convert irregular point clouds into ordered grid representations. Early works~\cite{simony2018complex,yang2018pixor} exploit a BEV image, obtained by projecting raw point clouds, which provides a compact 3D representation and preserves the scales of objects. Hand-crafted BEV representations are, however, limited to obtain accurate 3D features for object localization. Recent methods represent point clouds in a form of 3D voxels~\cite{du2020associate,he2020structure,lang2019pointpillars,ye2020hvnet,zhou2018voxelnet}. They often use 3D convolutions to extract features from the voxel representation, which is computationally inefficient. To alleviate this problem, PointPillars~\cite{lang2019pointpillars} propose to use a pseudo-image. Representing 3D voxels in a form of an image allows to exploit 2D convolutions, resulting in more efficient 3D object detection, compared to using 3D voxels directly. The voxel-based approaches mainly have two limitations: (1) Fine-grained 3D structures are lost during voxelization; (2) The localization performance largely depends on the size of voxel grids. As the grid size becomes smaller, fine-grained 3D representations can be obtained but at the cost of more runtime. Second, point-based approaches leverage raw point clouds directly. They extract point-based features, typically using PointNet++~\cite{qi2017pointnet,qi2017pointnet++}, that provide more accurate spatial representations than 3D voxels. These approaches thus perform better than the voxel-based one in terms of detection accuracy. It is, however, computationally demanding to extract point-based features directly from a large amount of point clouds for~\eg,~outdoor sceneries. To circumvent this issue, point-based approaches adopt a two-stage pipeline. Specifically, they estimate 3D object proposals either using point-based features~\cite{shi2019pointrcnn, yang2019std} or off-the-shelf 2D object detectors~\cite{qi2018frustum, wang2019frustum} in the first stage. The object proposals are then refined with the point-based features inside the proposals for final predictions in the second stage. Recently, PV-RCNN~\cite{shi2020pv} and Fast Point R-CNN~\cite{chen2019fast} further reduce the computational cost of point-based approaches. They propose to exploit voxel-based features to generate 3D object proposals in the first stage, while refining them to estimate final 3D bounding boxes with point-based features in the second stage. Since these methods exploit 3D voxels and raw point clouds separately in different stages, they do not fully leverage the complementary information of the two representations. Moreover, the point-based approaches are still slower than voxel-based methods, even leveraging a single-stage framework~\cite{yang20203dssd, wang2019frustum}.

      Our approach belongs to the voxel-based method in the sense that it uses a pseudo image representation of point clouds, similar to PointPillars~\cite{lang2019pointpillars}. In contrast to previous voxel-based methods, we also exploit point-based features for the pseudo image representation in an efficient way, which boosts the detection performance significantly, while retaining fast runtime. Our approach is similar to recent point-based methods~\cite{chen2019fast,shi2020pv,yang20203dssd} in that it exploits both voxel-based and point-based features~\cite{chen2019fast,shi2020pv}. On the contrary, we leverage both voxel-based and point-based representations jointly within a single-stage framework. We weave both representations in a form of a pseudo image, allowing to exploit 2D convolutions for efficient 3D object detection. Moreover, we further reduce the computational cost for extracting point-based features via augmenting a feature encoder with a memory module.  
      \vspace{-0.4cm}

 \paragraph{Memory networks.}
   Memory networks~\cite{weston2015memory} have been introduced to capture long-term dependencies in sequential data by augmenting a neural network with an external memory module that can be read and written to. Based on this idea, recent works develop more advanced architectures using continuous memory representations~\cite{sukhbaatar2015end} or a key-value memory~\cite{miller2016key}. A number of approaches have adopted the memory-augmented networks to solve various computer vision tasks including visual question answering~\cite{kumar2016ask, ma2018visual, fan2019heterogeneous}, one-shot learning~\cite{santoro2016meta,kaiser2017learning,cai2018memory}, anomaly detection~\cite{gong2019memorizing,park2020learning}, and person recognition~\cite{zhong2019invariance,xia2020online}.
   \vspace{-0.3cm}

\section{Approach}
\vspace{-0.2cm}
Our model mainly consists of three components~(Fig.~\ref{fig:overview}): a HVPR network~(Sec.~\ref{sec:pseudo-image}), a backbone network with AMFM~(Sec.~\ref{sec:backbone}), and a detection head~(Sec.~\ref{sec:head}). Given point clouds, the HVPR network outputs hybrid voxel-point representations in a form of pseudo-images. To this end, we exploit a two-stream encoder to extract voxel-based and point-based features. For each voxel-based feature, we aggregate point-based ones, based on their similarities, and obtain hybrid voxel-point representations. Extracting point-based features is, however, computationally demanding. To overcome this, we augment point-based features using a memory module. Specifically, we store various prototypes of point-based features in memory items, and aggregate the prototypical features in the memory to obtain voxel-memory representations. The memory items are updated by encouraging the aggregated prototypical and point-based features to be similar. Note that we exploit the voxel-memory representations only at test time, instead of using point-based ones directly, enabling a fast object detection. The backbone network inputs the voxel-memory representations in a form of a pseudo image, and extracts multi-scale feature maps. The AMFM refines the feature maps using 3D scale representations, and provides scale-aware features. The detection head predicts 3D object bounding boxes and object classes using the scale-aware features. In the following, we describe our single-stage detection framework in detail.
\vspace{-0.2cm}

\subsection{HVPR Network}\label{sec:pseudo-image}
\vspace{-0.2cm}
\paragraph{Voxel-based feature.}
 We assume that the physical dimension of a 3D scene that point clouds lie on is within the range of $W \times H \times L$, where $W$, $H$, and $L$ are width, height, and length, respectively, in three-dimensions. Supposing that the size of each voxel is $v_{W} \times v_{H} \times v_{L}$, we define voxel grids of size $W^\prime \times H^\prime \times L^\prime$, where $W^\prime=W/v_{W}$, $H^\prime=H/v_{H}$, $L^\prime=L/v_{L}$. Following~\cite{lang2019pointpillars}, we voxelize point clouds in a x-y plane only,~\ie,~by setting the value of $v_L$ to be the same as $L$. We represent a voxel as a tensor of size~$N_\mathrm{vox} \times D$, where $N_\mathrm{vox}$ and $D$ are the number of point clouds within each voxel and the size of augmented point clouds\footnote{We parameterize a point cloud using its positions at x, y, z coordinates, and a reflectance intensity of laser. Following~\cite{lang2019pointpillars}, we further augment each point cloud with average distances of all point clouds in the voxel and x-y offsets from the center position of the voxel, resulting in a $D$-dimensional feature vector.}, respectively. We exploit a tiny PointNet~\cite{qi2017pointnet} as an encoder for voxel-based features. The encoder takes a set of voxels and extracts features of all point clouds in each voxel, providing an output of size~$C \times N \times N_\mathrm{vox}$, where $N$ is a total number of voxels. It then applies a max pooling operator over the point clouds in each voxel to obtain a voxel-based feature map of size~$C \times N$, where we denote by $\mathbf{f}_\mathrm{vox}(n)$ each voxel-based feature of size~$C\times 1$, where $n=1,\cdots,N$.
 \vspace{-0.3cm}

\paragraph{Point-based feature.}
As a point stream, we adopt a PointNet++~\cite{qi2017pointnet++}, widely used to other 3D applications, such as 3D semantic segmentation~\cite{behley2019semantickitti,wang2019associatively} and 3D object detection~\cite{yang2019std,yang20203dssd, shi2019pointrcnn, qi2018frustum, wang2019frustum}, that extracts point-based features from raw point clouds directly. To this end, PointNet++ uses set abstraction~(SA) and feature propagation~(FP) layers. It first downsamples a set of points gradually, with a number of SA layers, to produce novel features with fewer points. The FP layers then propagate sub-sampled features to recover the initial points entirely. We follow these procedures to obtain a point-based feature for each point cloud. We denote by $\mathbf{f}_\mathrm{pts}(m)$ each point-based feature of size~$C\times 1$, where $m=1,\cdots,M$ and $M$ is a a total number of point clouds.

\vspace{-0.3cm}
\subsubsection{Voxel-point representation}\label{subsec:vp_representation}
\vspace{-0.2cm}
We integrate voxel-based and point-based features, $\mathbf{f}_\mathrm{vox}$ and $\mathbf{f}_\mathrm{pts}$, to obtain a hybrid voxel-point representation. To this end, we compute the dot product between all pairs of voxel-based and point-based features, resulting in a 2-dimensional correlation map of size~$N\times M$ as follows:
   \begin{equation}\label{eq:matching score}
      C(n,m) = \mathbf{f}_\mathrm{vox}(n)^\top\mathbf{f}_\mathrm{pts}(m).
   \end{equation}  
 For each voxel-based feature, we select the nearest $K$ point-based features, according to the correlation scores. We denote by~$\hat{\mathbf{f}}_\mathrm{pts}(n,k)$ the $k$-th similar point-wise feature w.r.t the $n$-th voxel-based one, where $k=1,\dots,K$. We then compute the matching probabilities between these voxel-based and point-based features as follows: 
 \begin{equation}\label{eq:v}
      P(n,k) = \frac{\exp(\mathbf{f}_\mathrm{vox}(n)^\top{\hat{\mathbf{f}}_\mathrm{pts}(n,k)})}{\sum_{k^\prime} \exp(\mathbf{f}_\mathrm{vox}(n)^\top{\hat{\mathbf{f}}_\mathrm{pts}(n,k^\prime)})}.
   \end{equation}
Finally, we aggregate the nearest $K$ point-based features with corresponding matching probabilities as follows:
\begin{equation}\label{eq:aggregated_point_feature}
	\mathbf{g}_\mathrm{pts}(n) = \sum_k P(n,k) \hat{\mathbf{f}}_\mathrm{pts}(n,k),
   \vspace{-0.2cm}
\end{equation} 
where $\mathbf{g}_\mathrm{pts}(n)$ is an aggregated point-based feature of size~$C\times 1$. It is semantically similar to the corresponding voxel-based one, but contains more accurate 3D structural information for objects. This enables imposing the fine-grained 3D representation to the voxel-based feature that loses the structure during voxelization. Specifically, we concatenate voxel-based and aggregated point-based features, and obtain a voxel-point pseudo image of size~$H \times W \times 2C$ by scattering the concatenated features back to the corresponding voxel locations, similar to PointPillars~\cite{lang2019pointpillars}. In this context, the aggregated feature also has an effect of feature augmentation, providing more discriminative 3D representations, especially for the voxels obtained from sparse point clouds. 
\vspace{-0.3cm}
\subsubsection{Voxel-memory representation}
\vspace{-0.2cm}
Point-wise features obtained by PointNet++~\cite{qi2017pointnet++} contain fine-grained 3D representations, but this requires lots of computational cost. PointRCNN~\cite{shi2019pointrcnn}, for instance, composed of 4 pairs of SA and FP layers, requires 54ms with a Titan V GPU to obtain point-based features~\cite{yang20203dssd}, which already exceeds the overall runtime of most voxel-based 3D detection methods~\cite{lang2019pointpillars,he2020structure,ye2020hvnet}.  To address this problem, we augment point-based features using a memory module, where each memory item stores  various prototypes of the features, to obtain voxel-memory representations.

Concretely, we denote by $\mathbf{f}_\mathrm{mem}(t)$ each item in the memory, the size of which is $C\times 1$, where $t=1,\cdots,T$, and $T$ is a total number of memory items. We integrate the voxel-based features~$\mathbf{f}_\mathrm{vox}$ and the memory items~$\mathbf{f}_\mathrm{mem}$ to obtain a voxel-memory pseudo image in a similar way to the voxel-point one in Sec.~\ref{subsec:vp_representation}. Specifically, we consider the memory as a set of prototypes for point-based features. To read memory items, we use voxel-based features as queries. We then compute matching probabilities between voxel-based features and memory items, and aggregate the top $K$ items with corresponding probabilities for each voxel-based feature, which is analogous to the aggregated point-based feature in~\eqref{eq:aggregated_point_feature}. We concatenate the voxel-based features with the aggregated items, and scatter the concatenated features back to the original voxel locations to form a voxel-memory pseudo image. To update memory items, we encourage the aggregated point-based features and memory items to be similar as follows:
 \begin{equation}\label{eq:memloss}
         \mathcal{L}_\mathrm{mem} = \sum_n \|\mathbf{g}_\mathrm{pts}(n)-\mathbf{g}_\mathrm{mem}(n)\|_2,
         \vspace{-0.2cm}
      \end{equation}
where we denote by $\mathbf{g}_\mathrm{mem}(n)$ an aggregated memory item, and $\Vert \cdot \Vert_{2}$ computes the L2 norm of a vector. Note that we do not use the voxel-point pseudo image at test time. We instead use the voxel-memory pseudo image only, avoiding extracting point-based features at test time. As will be shown in our experiments, obtaining hybrid 3D representations using a memory is much more cheaper than exploiting point-based features directly, while retaining the same level of quality.
\vspace{-0.2cm}   
\subsection{Backbone Network with AMFM}\label{sec:backbone}
\vspace{-0.2cm}
Our backbone network consists of a series of 2D convolutional layers and AMFM~(Fig.~\ref{fig:overview}). It inputs our hybrid 3D representations,~\ie,~the voxel-memory pseudo image, and gives multi-scale feature maps. The AMFM then refines them using spatial attention maps, and provides scale-aware features. We obtain final feature representations for predicting 3D object bounding boxes by concatenating the scale-aware feature maps along the channel dimension. 
\vspace{-0.5cm}

\begin{figure}[t]
   \captionsetup{font={small}}
      \centering
         \includegraphics[width=1.0\linewidth]{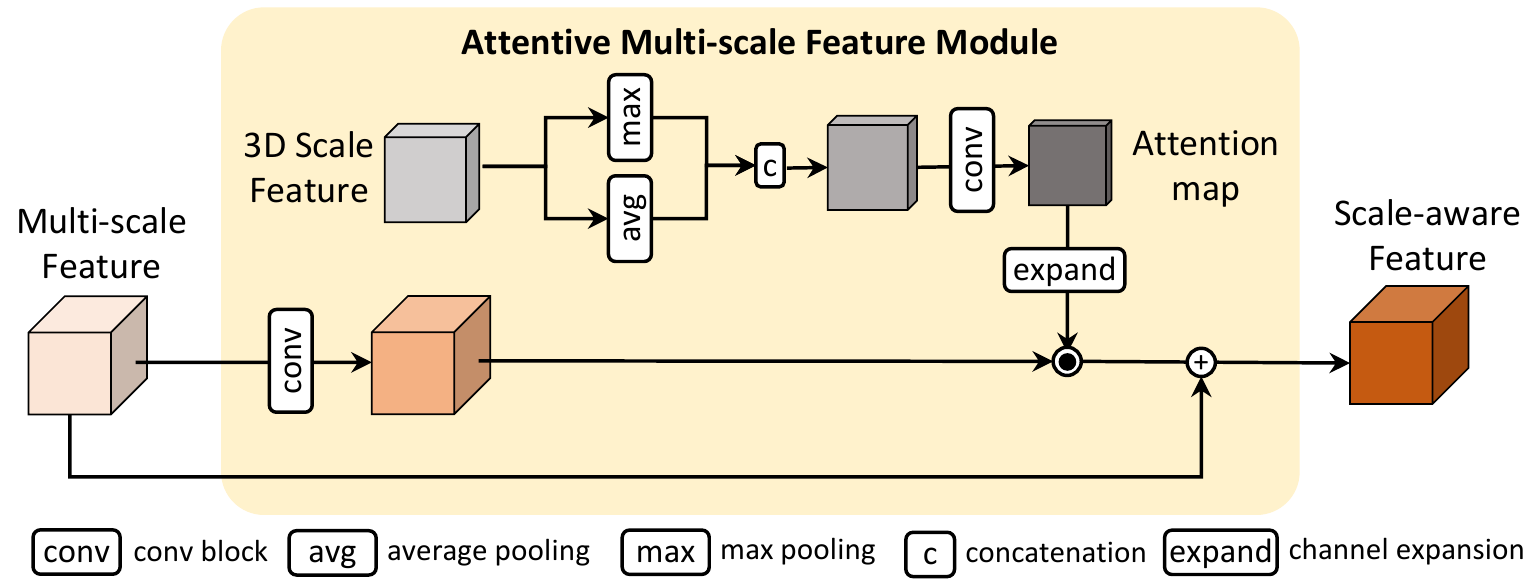}
\vspace{-0.5cm}
\caption{Illustration of AMFM. AMFM refines multi-scale features with a spatial attention map. It then combines the initial and refined features via a skip connection to obtain scale-aware representations. To compute the attention map, we leverage 3D scale features considering scale variations of objects in a 3D scene. See text for details.}

      \vspace{-0.6cm}
      \label{fig:AMFM}
    \end{figure}

  \paragraph{AMFM.} 
      There are large scale variations across 3D object instances in a scene. Multi-scale features help to consider the scale variations for 3D object detection, but they are not scale-aware, even when obtained from a feature pyramid~\cite{li2020netnet}. To address this problem, we introduce AMFM that provides scale-aware features using 3D scale information explicitly. The basic idea of AMFM is to exploit spatial attention maps~\cite{woo2018cbam}, while leveraging 3D scale features explicitly, for suppressing scale-confused features. Specifically, AMFM refines a multi-scale feature by element-wise multiplication with the attention map. It then combines the initial and refined multi-scale features via a skip connection. We illustrate in Fig.~\ref{fig:AMFM} how AMFM works in the backbone network in detail. In the following, we describe how to extract 3D scale features, and then present how to obtain a spatial attention map using 3D scale features in detail.

We observe that 3D point clouds are sparse with the densities largely varying w.r.t distances from a LiDAR sensor. Namely, the sparse and irregular patterns of point clouds and their distances from a sensor reflect scale information of 3D objects. Based on this idea, we represent each voxel with the number of point clouds within a voxel, and an absolute position of point clouds averaged on each voxel and its distance from a sensor. We input a set of voxel representations to a network, and obtain voxel-wise features. To be specific, we exploit a tiny PointNet~\cite{qi2017pointnet} similar to the encoder for voxel-based features. We then scatter the voxel-wise features back to the original locations to obtain a 3D scale feature map of the same size as the voxel-memory pseudo image. We downsample the 3D scale feature with a strided convolution, before feeding it to AMFM, such that it has the same spatial resolution as a corresponding multi-scale feature map. To obtain a spatial attention map, we apply max pooling and average pooling layers to the 3D scale feature along the channel dimension, highlighting informative regions in the feature map~\cite{komodakis2017paying}. We concatenate the pooled features, and feed the concatenated one to a convolutional layer, followed by applying a sigmoid function, to produce the attention map.

\vspace{-0.1cm}        
\subsection{Detection Head and Loss}\label{sec:head}
\vspace{-0.1cm}
Following~\cite{lang2019pointpillars}, we adopt the detection head of SSD~\cite{liu2016ssd} to localize 3D objects. It consists of two fully-connected layers for regressing 3D object bounding boxes and classifying object classes inside the boxes.  We parameterize a 3D object bounding box as $(x,y,z,w,l,h,\theta)$, where $(x,y,z)$, $(w,l,h)$ and $\theta$ are the center coordinate, the size, the heading angle of the box, respectively. We compute 2D intersection over union~(IoU) scores~\cite{everingham2010pascal} between anchor boxes and the ground truth, and pick the anchors with the IoU scores being larger than a predefined threshold. Our detection head regresses residuals between the anchor boxes and the ground truth directly, defined as follows:
\begin{equation}
\label{eq:residual}
            \begin{aligned}
               &\Delta x = \frac{x^\mathrm{gt}-x^\mathrm{a}}{d},
               \Delta y = \frac{y^\mathrm{gt}-y^\mathrm{a}}{d},
               \Delta z = \frac{z^\mathrm{gt}-z^\mathrm{a}}{h^\mathrm{a}} \\
               &\Delta w = \log \frac{w^\mathrm{gt}}{w^\mathrm{a}},
               \Delta l = \log \frac{l^\mathrm{gt}}{l^\mathrm{a}},
               \Delta h = \log \frac{h^\mathrm{gt}}{h^\mathrm{a}} \\
               &\Delta \theta = \sin{(\theta^\mathrm{gt}-\theta^\mathrm{a})},
            \end{aligned}            
         \end{equation}
where $d = \sqrt{\left(w^\mathrm{a}\right)^{2} + \left(l^\mathrm{a}\right)^{2}}$, and we use the superscripts, $\mathrm{gt}$ and $\mathrm{a}$, to indicate parameters for ground-truth and anchor boxes, respectively.

To train our model, we use four terms for regression~($\mathcal{L}_{\text{reg}}$ and $\mathcal{L}_{\text{dir}}$), classification~($\mathcal{L}_{\text{cls}}$), and memory update~($\mathcal{L}_{\text{mem}}$ in~\eqref{eq:memloss}), as follows: 
\begin{equation}
            \mathcal{L} = \frac{1}{N_{\mathrm{pos}}}(\lambda_{\mathrm{reg}}\mathcal{L_{\mathrm{reg}}}
                        + \lambda_{\mathrm{dir}}\mathcal{L}_{\mathrm{dir}}
            + \lambda_{\mathrm{cls}}\mathcal{L}_{\mathrm{cls}}
            + \lambda_{\mathrm{mem}}\mathcal{L}_{\mathrm{mem}}),
         \end{equation}
         where $N_{\mathrm{pos}}$ is the number of positive anchors, and $\lambda$ is a balancing parameter for the corresponding loss. We define the regression term with the residual between anchor and ground-truth boxes as follows:
          \begin{equation}
            \mathcal{L}_{\text{reg}} = \sum_{r \in (x,y,z,w,l,h,\theta)} \text{SmoothL1}(\Delta r).
            \end{equation}
            Note that $\Delta \theta$ is defined with a sine function in~\eqref{eq:residual}, penalizing the same 3D bounding boxes but with opposite directions differently. To prevent this, we also exploit the angle classification term~($\mathcal{L}_{\text{dir}}$)~\cite{yan2018second}. As a classification term~($\mathcal{L}_{\text{cls}}$), we use the focal loss~\cite{lin2017focal}, which has shown the effectiveness to handle a class imbalance  between positive and negative samples, with the default parameters in~\cite{lin2017focal}.

\vspace{-0.2cm}
\section{Experiments}
\vspace{-0.1cm}
\subsection{Implementation details}
\vspace{-0.1cm}
\paragraph{Dataset and evaluation protocol.}
We evaluate our model on the KITTI 3D object detection benchmark~\cite{geiger2012we}. The KITTI dataset provides 7,481 training and 7,518 test samples for object classes of cars, pedestrians and cyclists. Object instances across different classes are further classified into easy, moderate and hard splits, depending on the object size, the degree of occlusion and the maximum truncation level. We evaluate our model on the car and the pedestrian classes, most commonly used object class and the hardest one for 3D object detection, respectively. Following~\cite{liu2020tanet}, we split the training set into two splits with the ratio of around 5:1 to train our model for testing via an official evaluation server. For other experiments including an ablation study, we follow the experimental protocol in~\cite{chen20153d,everingham2010pascal}, and split the original training set of 7,481 samples into 3,712 and 3,769 ones for training and validation, respectively. We report the mean average precision~(mAP) on both validation and test splits with 40 recall positions on a PR curve~\cite{geiger2012we}. At test time, we apply non maximum suppression (NMS) with a threshold of 0.1 to eliminate multiple overlapping bounding boxes.
\vspace{-0.5cm}

\paragraph{Training.} 
Following the previous works~\cite{lang2019pointpillars,yan2018second,zhou2018voxelnet,liu2020tanet}, we train separate networks for the car and the pedestrian classes. We describe hereafter the training setting for the car class, and the one for the pedestrian is included in the supplementary material. We train our network for 100 epochs, with a learning rate of 3e-3 and a weight decay of 1e-2. We use a cosine annealing technique~\cite{loshchilov2016sgdr} as a learning rate scheduler. Batch size is set to 8 for the network. All models are trained end-to-end using \texttt{PyTorch}~\cite{paszke2017automatic}. We use data augmentation techniques during training. Specifically, we apply random flipping along the $x$-axis, global scaling, where a scaling factor is randomly chosen within a range of $[0.95, 1.05]$, and global rotation along the $z$-axis, where a rotation angle is sampled randomly from a range of $[-\frac{\pi}{4}, \frac{\pi}{4}]$. We also use a ground-truth augmentation technique~\cite{yan2018second} that randomly selects ground-truth bounding boxes within the entire training dataset and associates them for each 3D scene. 
\vspace{-0.5cm}

\paragraph{Parameter setting.}
We set the parameters for our model by a grid search in terms of mAP on the validation split: $K=20$, $C=64$, $T=2,000$, $\lambda_{\mathrm{reg}}= 2.0$, $\lambda_{\mathrm{dir}}= 0.2$, $\lambda_{\mathrm{cls}}=1.0$ and $\lambda_{\mathrm{mem}}=1.0$. For other parameters, we use the same setting as in~\cite{lang2019pointpillars} as follows: We assume that the dimension of a 3D scene~($W,H,L$) is within a range of $[(0,70.4), (-40,40), (-3,1)]$ meters. We set the size of anchors as $1.6\times 3.9\times 1.5$. To define positive and negative pairs, we compute IoU scores between the anchors and the ground truth. We choose the anchors with the IoU scores being larger than 0.6, as positive boxes, while those lower than 0.45 are used as negative ones. We use $(0.16, 0.16, 4)$ as the size of a voxel,~$v_W \times v_H \times v_L$. The number of point clouds within each voxel~$N_\mathrm{vox}$ and the size of augmented point clouds~$D$ are set to 32 and 9, respectively. Please refer to the supplementary material for more implementation details including our network architecture.
\vspace{-0.3cm}

\setlength{\tabcolsep}{0.2em}
            \begin{table*}[t]
            \captionsetup{font={small}}
            \small
            \begin{center}
            \caption{Quantitative comparison with the state of the art in terms of mAP(\%) and runtime on the KITTI test set~\cite{geiger2012we}. We mainly compare our model with voxel-based methods using pseudo image representations~(Voxel-PI). For comparison, we also report results for point-based and~(Point) other voxel-based methods using 3D voxel representations~(Voxel-3D), although they typically provide better results, but at the cost of much slower runtime, than exploiting pseudo image representations. Numbers in bold for Voxel-PI indicate the best performance and underscored ones are the second best.}
               \vspace{-0.3cm}
               \begin{tabular}{l c c c c c c c c c c c c c }
                  \hline
                  \multicolumn{2}{c}{\multirow{2}{*}{\parbox{3em}{\centering Type}}}& \multirow{2}{*}{\parbox{3em}{\centering Stage}}&\multirow{2}{*}{\parbox{5em}{\centering Methods}} &\multirow{2}{*}{\parbox{5em}{\centering Reference}}&\multirow{2}{*}{\parbox{6em}{\centering Runtime (Hz)}}  & \multirow{2}{*}{\parbox{4em}{\centering GPU}} & \multicolumn{3}{c}{\parbox{9em}{\centering Car}} &\multicolumn{3}{c}{\parbox{9em}{\centering Pedestrian}} \\
                  &&&&&&  &\parbox{3em}{\centering Easy}& \parbox{3em}{\centering Mod.} & \parbox{3em}{\centering Hard} &\parbox{3em}{\centering Easy}& \parbox{3em}{\centering Mod.} & \parbox{3em}{\centering Hard}    \\
                  \cmidrule{1-13}\morecmidrules\hline
                  \multicolumn{2}{c}{\multirow{6}{*}{Point~}} 
                  &Two&PointRCNN~\cite{shi2019pointrcnn} &CVPR 2019&10 &TITAN XP&86.96  &75.64  &70.70 &47.98  &39.37  &36.01     \\
                  &&Two&FastPointRCNN~\cite{chen2019fast}&ICCV 2019&16.7 &Tesla P40 &85.29  &77.40  &70.24 &-  &-  &-  \\
                  &&One&F-ConvNet~\cite{wang2019frustum}&IROS 2019&2.1 &GTX 1080 &87.36  &76.39  &66.69 &52.16  &43.38  &38.80    \\
                  &&Two&STD~\cite{yang2019std} &ICCV 2019&12.5  &TITAN V&87.95  &79.71  &75.09 &53.29 &42.47  &38.35     \\
                  &&One&3DSSD~\cite{yang20203dssd} &CVPR 2020&25  &TITAN V& 88.36 &79.57  &74.55 &54.64  &44.27  &40.23   \\
                  &&Two&PV-RCNN~\cite{shi2020pv} &CVPR2020& 12.5 &GTX 1080Ti&90.25  &81.43  &76.82 &52.17  &43.29  &40.29    \\
                  \midrule
                  \multicolumn{2}{c}{\multirow{4}{*}{Voxel-3D~}}
                  &One&VoxelNet~\cite{zhou2018voxelnet}&CVPR 2018& 4.5&TITAN X& 77.47 &65.11  &57.73 &39.48  &33.69  &31.51   \\
                  &&One&3DIoULoss~\cite{zhou2019iou}&3DV 2019&12.5 & - & 86.16 & 76.50  &71.39 &-  &-  &-     \\
                  &&One&SA-SSD~\cite{he2020structure} &CVPR 2020&25 &GTX 2080Ti&88.75  &79.79  & 74.16 &-  &-  &-    \\
                  &&One&HotSpotNet~\cite{chen2019object} &ECCV 2020&25 &-&87.60 &78.31  &73.34 &53.10  &45.37  &41.47   \\
                  \midrule
                  \multicolumn{2}{c}{\multirow{5}{*}{Voxel-PI~}}
                  &One&SECOND~\cite{yan2018second} &Sensors 2018& 20&GTX 1080Ti& 84.78 & 75.32 &68.70 &45.31  &35.52  &33.14    \\
                  &&One&PointPillars~\cite{lang2019pointpillars}&CVPR 2019& \textbf{42.4}&GTX 1080Ti& 82.58 &74.31  &68.99 &51.45  &41.92  &38.89 \\
                  &&One&TANet~\cite{liu2020tanet} &AAAI 2020& 28.5&TITAN V& 84.39& 75.94 & 68.82 &\textbf{53.72}  &\textbf{44.34}  &\underline{40.49}     \\
                  &&One&Associate-3D~\cite{du2020associate} &CVPR 2020&20&GTX 1080Ti& \underline{85.99} & \underline{77.40} & \underline{70.53} &-  &-  &-  \\
                  &&One&Ours& & \underline{36.1}  &GTX 2080Ti&\textbf{86.38}  & \textbf{77.92} & \textbf{73.04} &\underline{53.47}  &\underline{43.96}  &\textbf{40.64}   \\
                  \hline
               \end{tabular}
               \label{table:Comparison}
            \end{center}	
               \vspace{-0.7cm}
         \end{table*}
         
\subsection{Results}
\vspace{-0.2cm}
We compare in Table~\ref{table:Comparison} our model with the state of the art for LiDAR based 3D detection. We report mAP scores of car and pedestrian classes on the KITTI test dataset~\cite{geiger2012we}. All numbers including ours are obtained from an official evaluation server, except for~\cite{zhou2018voxelnet,yan2018second}, which are taken from~\cite{chen2019object}. We classify 3D object detection methods into point-based and voxel-based approaches, according to features used to represent 3D objects. The voxel-based approaches are further categorized into two groups based on whether they exploit 3D voxels explicitly or implicitly in a form of a pseudo image, where our model belongs to the latter one. We mainly compare our model with voxel-based methods, especially with the ones exploiting pseudo image representations. For comparison, we also report results of point-based and other voxel-based approaches. From this table, we observe four things: (1) Our model for the car class gives the best results among voxel-based methods using pseudo images. This demonstrates the effectiveness of our approach to exploiting a hybrid 3D representation and scale-aware features for 3D object detection. (2) Our method is fastest among voxel-based approaches, except PointPillars~\cite{lang2019pointpillars}, while providing competitive results. Although PointPillars is faster than ours~(42.4Hz vs. 36.1Hz),  it is outperformed by our model by a significant margin in terms of mAP for all difficulty levels. This suggests that our model offers a good compromise in terms of speed and accuracy. (3) The performance improvements over other voxel-based methods using pseudo images are particularly significant in a hard split, where 3D objects are typically small in size and largely occluded. For example, the mAP gain of our model w.r.t the second best method~\cite{du2020associate} (73.04 vs. 70.53) is much larger than those for easy and moderate splits of the car class. Also, our model gives the best result among other voxel-based methods using pseudo images on the hard split of the pedestrian class. This confirms that augmenting point-based features provides more discriminative 3D representations, especially for objects taken from sparse point clouds. (4) Our model for the pedestrian class outperforms PointPillars~\cite{lang2019pointpillars} significantly, while providing a competitive result with the best voxel-based method using pseudo images~\cite{liu2020tanet}. This shows that our model can generalize to other object classes. Note that our 3D representations from the HVPR network can be easily incorporated into other voxel-based methods~\cite{zhou2018voxelnet,he2020structure,zhou2019iou}, which may boost the mAP performance.

\begin{figure*}
   \captionsetup{font={small}}
   \centering
         \includegraphics[width=1.0\linewidth]{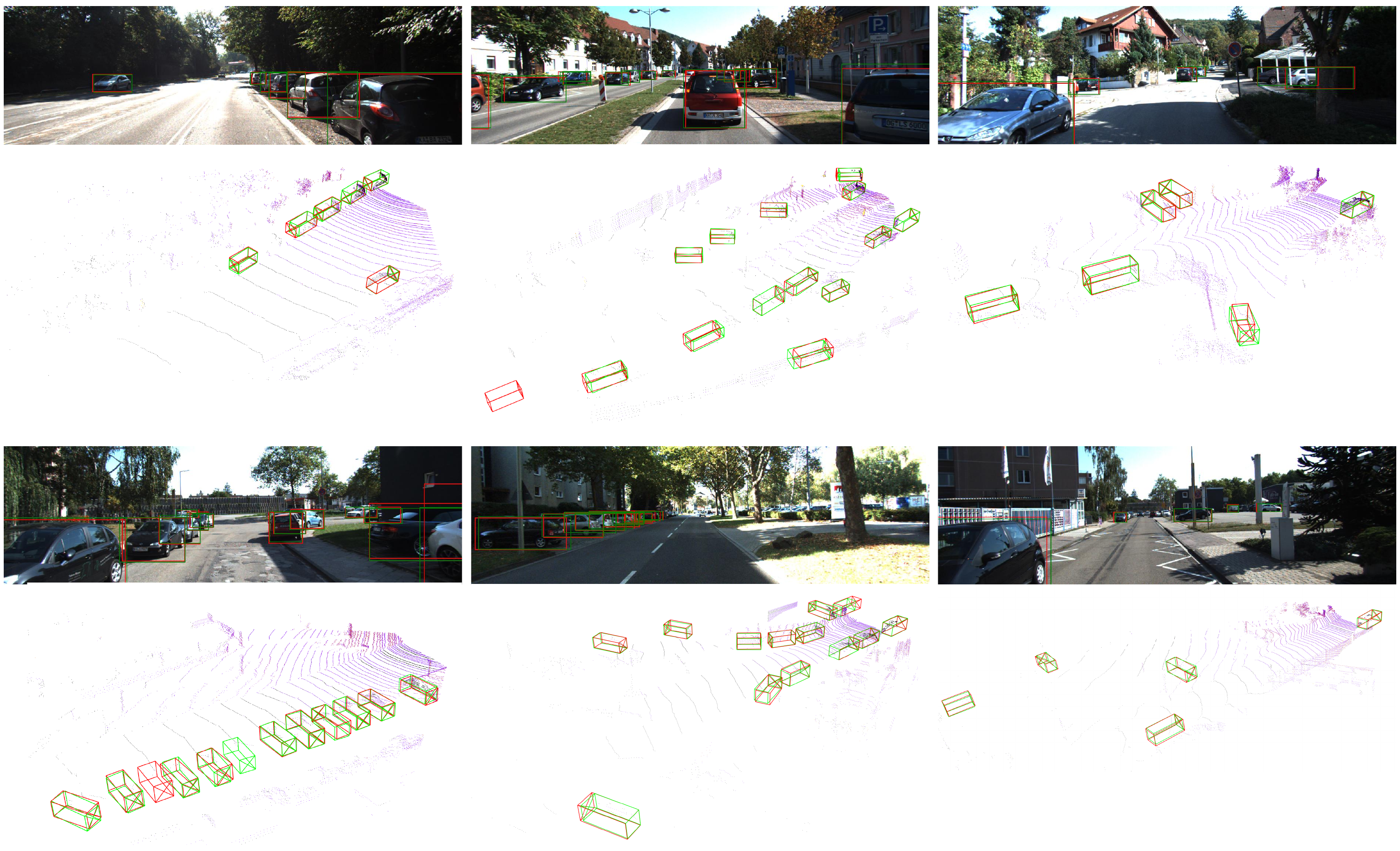}
      \vspace{-0.6cm}
      \caption{Qualitative results on the validation split of KITTI~\cite{geiger2012we}. Our predictions and ground-truth bounding boxes are shown in green and red, respectively. Our method localizes small and/or occluded objects well, except the heavily occluded ones,~\eg,~in the left-bottom of the top-middle image. We also show 2D bounding boxes projected from 3D detection results. Best viewed in color. }
      \vspace{-0.4cm}
      \label{fig:quali}
      \end{figure*}
      
\setlength{\tabcolsep}{0.3em}
\begin{table}
\captionsetup{font={small}}
\small
   \begin{center}
      \caption{Quantitative comparison for variants of our model. We compute mAP~(\%) and measure runtime~(Hz) on the KITTI validation set~\cite{geiger2012we}. VoxF: Voxel-based feature; PtsF: Point-based feature; MEM: Memory module; 3DSF: 3D scale feature.}
   \vspace{-0.3cm}
      \label{table:Ablation}
      \begin{tabular}{c c c c c c c c c} 
         \hline
         {\centering VoxF}	& {\centering PtsF} & {\centering Mem}& {\centering AMFM} & {\centering  3DSF} & {\centering Easy} & {\centering Mod.} & {\centering Hard} & {\centering Time }	\\
		\hline
          \cmark		&			&			&	 &	&	87.66 &	78.82	& 75.32 & 	39.5\\
          \cmark		&	\cmark		&		&	 &	 &	88.97 &	80.35	&77.28 & 	6.8	\\
          \cmark	   &	\cmark		&	\cmark		&	 &	 &	87.96 &	80.44	&77.92 & 37.9	\\
          \cmark		&	\cmark		&	\cmark		&	\cmark &	 &	89.51 &  80.86	& 78.22 & 31.3	\\
          \cmark	   &	\cmark		&	\cmark		&\cmark	 &	\cmark &	\bf{91.14} &	\bf{82.05}	& \bf{79.49} & 	36.1	\\
         \hline
      \end{tabular}
   \end{center}
   \vspace{-0.7cm}
\end{table}

\subsection{Discussion}
\vspace{-0.1cm}
In this section, we show more experimental results to analyze individual components of our model in detail. Following previous works~\cite{zhou2018voxelnet,yang2019std,yang20203dssd}, we report quantitative results of the car class in terms of mAP on the validation split of KITTI~\cite{geiger2012we}. We measure the mAP performance with 40 recall positions.

\vspace{-0.5cm}
\paragraph{Ablation study.}
We show an ablation analysis on different components of our model in Table~\ref{table:Ablation}. As our baseline in the first row, we use PointPillars~\cite{lang2019pointpillars}, reproduced by ourselves, for ablation studies. Similar to ours, PointPillars leverage a pseudo image to represent a 3D scene, but it exploits voxel-based features alone. We can see from the first and second rows that a hybrid representation using additional point-based features boosts the detection performance remarkably, which provides mAP gains of 1.31, 1.53 and 1.94 for easy, moderate and hard splits, respectively. The improvements for the moderate and hard splits are more significant than the easy one. This suggests that 1)~the aggregated point-based features allow to obtain more discriminative 3D representations, especially for small or occluded objects with sparse points, and 2) voxel-based and point-based features are complementary to each other. Extracting point-based features, however, is computationally expensive, which degrades runtime performance from 39.5Hz to 6.8Hz. The third row demonstrates that our memory module boosts the runtime performance significantly, while retaining the same level of quality. The average runtime is almost the same as the baseline~(39.5Hz vs. 37.9Hz). This shows the effectiveness and efficiency of a memory module to augment point-based features. Our model in the fourth row exploits AMFM but without using 3D scale features. That is, it computes an attention map directly using multi-scale features. From the third and fourth rows, we can see that refining multi-scale features with a spatial attention map gives better results. The last two rows show that 3D scale features further boosts the detection performance drastically for all cases. The mAP gains for each split are 1.63, 1.19 and 1.27, respectively. The reason is that the features allow to leverage 3D scale information explicitly, which is particularly important to consider complex scale variations of objects in a scene for 3D object detection.

\begin{table}
   \captionsetup{font={small}}
   \small
      \begin{center}
         \caption{Quantitative comparison for different combinations of a total numbers of memory items~($T$) and the number of nearest items retrieved~($K$). We report mAP~(\%) for the car class on the validation split of KITTI~\cite{geiger2012we}.}
      \vspace{-0.3cm}
         \label{table:memk}
         \begin{tabular}{c | c c c c }
             &  {\parbox{4em}{\centering $K=5$}}& {\parbox{4em}{\centering $K=10$}} & {\parbox{4em}{\centering $K=20$}} &{\parbox{4em}{\centering $K=30$}} \\
            \hline
             $T=1000$	 &	80.22	&	80.74		&	81.32		& 80.82	 	\\
            $T=2000$	  &	80.68	&	80.91		&	82.05	& 81.18	 	\\
            $T=3000$	 &	80.68 & 80.82		&	81.56		& 81.45	  	\\
            \hline
         \end{tabular}
      \end{center}
      \vspace{-0.7cm}
   \end{table}

\vspace{-0.5cm}
\paragraph{Memory items.}
We show in Table~\ref{table:memk} quantitative results for different combinations of a total number of memory items~$T$ and the number of nearest items retrieved~$K$. We compare the performance of our final model, with a memory module and AMFM, in terms of mAP, and report results of the car class on the moderate split. We can clearly see that the mAP performance is robust over the total number of memory items. We can also observe that aggregating more items for hybrid 3D representations provides better results, but using less similar items are not helpful. For example, mAP scores increase, until we reach the number of nearest items~$K$ to the value of 20. After that, using more items rather lowers the scores.

\vspace{-0.5cm}
\paragraph{Qualitative results.}
We visualize in Fig.~\ref{fig:quali} detection results on the validation split of KITTI~\cite{geiger2012we}. We can see that our model is robust to complex scale variations. That is, it localizes 3D objects well, including small ones captured with sparse point clouds. This confirms the effectiveness of our hybrid 3D representations and scale-aware features once again. More qualitative results are available in the supplementary material.
\vspace{-0.2cm}

\section{Conclusion}
\vspace{-0.2cm}
We have introduced a novel single-stage 3D detection method using hybrid 3D representations. We have integrated voxel-based and point-based features effectively to obtain 3D representations of a scene in a form of a pseudo image. For efficient object detection, we have proposed to use a memory module to augment point-based features. We have also presented AMFM that provides scale-aware representations using 3D scale features, which boosts the performance of 3D object detection significantly. An extensive experimental analysis on KITTI demonstrates that our model achieves the best results among voxel-based methods using pseudo images, and it provides a better compromise in terms of accuracy and speed than other 3D detection approaches.
\vspace{-0.5cm}

{\small
\paragraph{Acknowledgments.}This research was partly supported by R\&D program for Advanced Integrated-intelligence for Identification~(AIID) through the National Research Foundation of KOREA~(NRF) funded by Ministry of Science and ICT~(NRF-2018M3E3A1057289), and Institute for Information and Communications Technology Promotion~(IITP) funded by the Korean Government~(MSIP) under Grant 2016-0-00197.
}

{\small
\bibliographystyle{ieee_fullname}
\bibliography{egbib}

\begin{thebibliography}{10}\itemsep=-1pt

\bibitem{behley2019semantickitti}
Jens Behley, Martin Garbade, Andres Milioto, Jan Quenzel, Sven Behnke, Cyrill
  Stachniss, and Jurgen Gall.
\newblock Semantic{K}{I}{T}{T}{I}: A dataset for semantic scene understanding
  of lidar sequences.
\newblock In {\em ICCV}, 2019.

\bibitem{cai2018memory}
Qi Cai, Yingwei Pan, Ting Yao, Chenggang Yan, and Tao Mei.
\newblock Memory matching networks for one-shot image recognition.
\newblock In {\em CVPR}, 2018.

\bibitem{chen2019object}
Qi Chen, Lin Sun, Zhixin Wang, Kui Jia, and Alan Yuille.
\newblock Object as hotspots: An anchor-free 3{D} object detection approach via
  firing of hotspots.
\newblock In {\em ECCV}, 2020.

\bibitem{chen20153d}
Xiaozhi Chen, Kaustav Kundu, Yukun Zhu, Andrew~G Berneshawi, Huimin Ma, Sanja
  Fidler, and Raquel Urtasun.
\newblock 3{D} object proposals for accurate object class detection.
\newblock In {\em NeurIPS}, 2015.

\bibitem{chen2017multi}
Xiaozhi Chen, Huimin Ma, Ji Wan, Bo Li, and Tian Xia.
\newblock Multi-view 3{D} object detection network for autonomous driving.
\newblock In {\em CVPR}, 2017.

\bibitem{chen2019fast}
Yilun Chen, Shu Liu, Xiaoyong Shen, and Jiaya Jia.
\newblock Fast point r-cnn.
\newblock In {\em ICCV}, 2019.

\bibitem{du2020associate}
Liang Du, Xiaoqing Ye, Xiao Tan, Jianfeng Feng, Zhenbo Xu, Errui Ding, and
  Shilei Wen.
\newblock Associate-3{D}det: Perceptual-to-conceptual association for 3{D}
  point cloud object detection.
\newblock In {\em CVPR}, 2020.

\bibitem{everingham2010pascal}
Mark Everingham, Luc Van~Gool, Christopher~KI Williams, John Winn, and Andrew
  Zisserman.
\newblock The pascal visual object classes ({V}{O}{C}) challenge.
\newblock {\em IJCV}, 88, 2010.

\bibitem{fan2019heterogeneous}
Chenyou Fan, Xiaofan Zhang, Shu Zhang, Wensheng Wang, Chi Zhang, and Heng
  Huang.
\newblock Heterogeneous memory enhanced multimodal attention model for video
  question answering.
\newblock In {\em CVPR}, 2019.

\bibitem{geiger2012we}
Andreas Geiger, Philip Lenz, and Raquel Urtasun.
\newblock Are we ready for autonomous driving? the {K}{I}{T}{T}{I} vision
  benchmark suite.
\newblock In {\em CVPR}, 2012.

\bibitem{gong2019memorizing}
Dong Gong, Lingqiao Liu, Vuong Le, Budhaditya Saha, Moussa~Reda Mansour, Svetha
  Venkatesh, and Anton van~den Hengel.
\newblock Memorizing normality to detect anomaly: Memory-augmented deep
  autoencoder for unsupervised anomaly detection.
\newblock In {\em ICCV}, 2019.

\bibitem{he2020structure}
Chenhang He, Hui Zeng, Jianqiang Huang, Xian-Sheng Hua, and Lei Zhang.
\newblock Structure aware single-stage 3{D} object detection from point cloud.
\newblock In {\em CVPR}, 2020.

\bibitem{huang2020epnet}
Tengteng Huang, Zhe Liu, Xiwu Chen, and Xiang Bai.
\newblock E{P}{N}et: Enhancing point features with image semantics for 3{D}
  object detection.
\newblock In {\em ECCV}, 2020.

\bibitem{kaiser2017learning}
{\L}ukasz Kaiser, Ofir Nachum, Aurko Roy, and Samy Bengio.
\newblock Learning to remember rare events.
\newblock In {\em ICLR}, 2017.

\bibitem{komodakis2017paying}
Nikos Komodakis and Sergey Zagoruyko.
\newblock Paying more attention to attention: improving the performance of
  convolutional neural networks via attention transfer.
\newblock In {\em ICLR}, 2017.

\bibitem{ku2018joint}
Jason Ku, Melissa Mozifian, Jungwook Lee, Ali Harakeh, and Steven~L Waslander.
\newblock Joint 3{D} proposal generation and object detection from view
  aggregation.
\newblock In {\em IROS}, 2018.

\bibitem{kumar2016ask}
Ankit Kumar, Ozan Irsoy, Peter Ondruska, Mohit Iyyer, James Bradbury, Ishaan
  Gulrajani, Victor Zhong, Romain Paulus, and Richard Socher.
\newblock Ask me anything: Dynamic memory networks for natural language
  processing.
\newblock In {\em ICML}, 2016.

\bibitem{lang2019pointpillars}
Alex~H Lang, Sourabh Vora, Holger Caesar, Lubing Zhou, Jiong Yang, and Oscar
  Beijbom.
\newblock Pointpillars: Fast encoders for object detection from point clouds.
\newblock In {\em CVPR}, 2019.

\bibitem{li2020netnet}
Yazhao Li, Yanwei Pang, Jianbing Shen, Jiale Cao, and Ling Shao.
\newblock {N}{E}{T}{N}et: Neighbor erasing and transferring network for better
  single shot object detection.
\newblock In {\em CVPR}, 2020.

\bibitem{liang2019multi}
Ming Liang, Bin Yang, Yun Chen, Rui Hu, and Raquel Urtasun.
\newblock Multi-task multi-sensor fusion for 3{D} object detection.
\newblock In {\em CVPR}, 2019.

\bibitem{lin2017focal}
Tsung-Yi Lin, Priya Goyal, Ross Girshick, Kaiming He, and Piotr Doll{\'a}r.
\newblock Focal loss for dense object detection.
\newblock In {\em ICCV}, 2017.

\bibitem{liu2016ssd}
Wei Liu, Dragomir Anguelov, Dumitru Erhan, Christian Szegedy, Scott Reed,
  Cheng-Yang Fu, and Alexander~C Berg.
\newblock {S}{S}{D}: Single shot multibox detector.
\newblock In {\em ECCV}. Springer, 2016.

\bibitem{liu2020tanet}
Zhe Liu, Xin Zhao, Tengteng Huang, Ruolan Hu, Yu Zhou, and Xiang Bai.
\newblock {T}{A}{N}et: Robust 3{D} object detection from point clouds with
  triple attention.
\newblock In {\em AAAI}, 2020.

\bibitem{loshchilov2016sgdr}
Ilya Loshchilov and Frank Hutter.
\newblock {S}{G}{D}{R}: Stochastic gradient descent with warm restarts.
\newblock {\em arXiv preprint arXiv:1608.03983}, 2016.

\bibitem{ma2018visual}
Chao Ma, Chunhua Shen, Anthony Dick, Qi Wu, Peng Wang, Anton van~den Hengel,
  and Ian Reid.
\newblock Visual question answering with memory-augmented networks.
\newblock In {\em CVPR}, 2018.

\bibitem{miller2016key}
Alexander Miller, Adam Fisch, Jesse Dodge, Amir-Hossein Karimi, Antoine Bordes,
  and Jason Weston.
\newblock Key-value memory networks for directly reading documents.
\newblock In {\em EMNLP}, 2016.

\bibitem{park2020learning}
Hyunjong Park, Jongyoun Noh, and Bumsub Ham.
\newblock Learning memory-guided normality for anomaly detection.
\newblock In {\em CVPR}, 2020.

\bibitem{paszke2017automatic}
Adam Paszke, Sam Gross, Soumith Chintala, Gregory Chanan, Edward Yang, Zachary
  DeVito, Zeming Lin, Alban Desmaison, Luca Antiga, and Adam Lerer.
\newblock Automatic differentiation in pytorch.
\newblock 2017.

\bibitem{qi2018frustum}
Charles~R Qi, Wei Liu, Chenxia Wu, Hao Su, and Leonidas~J Guibas.
\newblock Frustum pointnets for 3{D} object detection from rgb-d data.
\newblock In {\em CVPR}, 2018.

\bibitem{qi2017pointnet}
Charles~R Qi, Hao Su, Kaichun Mo, and Leonidas~J Guibas.
\newblock Pointnet: Deep learning on point sets for 3{D} classification and
  segmentation.
\newblock In {\em CVPR}, 2017.

\bibitem{qi2017pointnet++}
Charles~Ruizhongtai Qi, Li Yi, Hao Su, and Leonidas~J Guibas.
\newblock Pointnet++: Deep hierarchical feature learning on point sets in a
  metric space.
\newblock In {\em NeurIPS}, 2017.

\bibitem{santoro2016meta}
Adam Santoro, Sergey Bartunov, Matthew Botvinick, Daan Wierstra, and Timothy
  Lillicrap.
\newblock Meta-learning with memory-augmented neural networks.
\newblock In {\em ICML}, 2016.

\bibitem{shi2020pv}
Shaoshuai Shi, Chaoxu Guo, Li Jiang, Zhe Wang, Jianping Shi, Xiaogang Wang, and
  Hongsheng Li.
\newblock {P}{V}-{R}{C}{N}{N}: Point-voxel feature set abstraction for 3{D}
  object detection.
\newblock In {\em CVPR}, 2020.

\bibitem{shi2019pointrcnn}
Shaoshuai Shi, Xiaogang Wang, and Hongsheng Li.
\newblock Point{R}{C}{N}{N}: 3{D} object proposal generation and detection from
  point cloud.
\newblock In {\em CVPR}, 2019.

\bibitem{simony2018complex}
Martin Simon, Stefan Milzy, Karl Amendey, and Horst-Michael Gross.
\newblock Complex-{Y}{O}{L}{O}: An euler-region-proposal for real-time 3d
  object detection on point clouds.
\newblock In {\em ECCVW}, 2018.

\bibitem{sukhbaatar2015end}
Sainbayar Sukhbaatar, Jason Weston, Rob Fergus, et~al.
\newblock End-to-end memory networks.
\newblock In {\em NeurIPS}, 2015.

\bibitem{wang2019associatively}
Xinlong Wang, Shu Liu, Xiaoyong Shen, Chunhua Shen, and Jiaya Jia.
\newblock Associatively segmenting instances and semantics in point clouds.
\newblock In {\em CVPR}, 2019.

\bibitem{wang2019frustum}
Zhixin Wang and Kui Jia.
\newblock Frustum convnet: Sliding frustums to aggregate local point-wise
  features for amodal 3{D} object detection.
\newblock In {\em IROS}, 2019.

\bibitem{weston2015memory}
Jason Weston, Sumit Chopra, and Antoine Bordes.
\newblock Memory networks.
\newblock In {\em ICLR}, 2015.

\bibitem{woo2018cbam}
Sanghyun Woo, Jongchan Park, Joon-Young Lee, and In So~Kweon.
\newblock {C}{B}{A}{M}: Convolutional block attention module.
\newblock In {\em ECCV}, 2018.

\bibitem{xia2020online}
Jiangyue Xia, Anyi Rao, Qingqiu Huang, Linning Xu, Jiangtao Wen, and Dahua Lin.
\newblock Online multi-modal person search in videos.
\newblock In {\em ECCV}, 2020.

\bibitem{yan2018second}
Yan Yan, Yuxing Mao, and Bo Li.
\newblock {S}{E}{C}{O}{N}{D}: Sparsely embedded convolutional detection.
\newblock {\em Sensors}, 18(10), 2018.

\bibitem{yang2018pixor}
Bin Yang, Wenjie Luo, and Raquel Urtasun.
\newblock {P}{I}{X}{O}{R}: Real-time 3{D} object detection from point clouds.
\newblock In {\em CVPR}, 2018.

\bibitem{yang20203dssd}
Zetong Yang, Yanan Sun, Shu Liu, and Jiaya Jia.
\newblock 3{D}{S}{S}{D}: Point-based 3{D} single stage object detector.
\newblock In {\em CVPR}, 2020.

\bibitem{yang2019std}
Zetong Yang, Yanan Sun, Shu Liu, Xiaoyong Shen, and Jiaya Jia.
\newblock S{T}{D}: Sparse-to-dense 3{D} object detector for point cloud.
\newblock In {\em ICCV}, 2019.

\bibitem{ye2020hvnet}
Maosheng Ye, Shuangjie Xu, and Tongyi Cao.
\newblock {H}{V}{N}et: Hybrid voxel network for lidar based 3{D} object
  detection.
\newblock In {\em CVPR}, 2020.

\bibitem{yoo20203d}
Jin~Hyeok Yoo, Yeocheol Kim, Ji~Song Kim, and Jun~Won Choi.
\newblock 3{D}-{C}{V}{F}: Generating joint camera and lidar features using
  cross-view spatial feature fusion for 3{D} object detection.
\newblock In {\em ECCV}, 2020.

\bibitem{zhong2019invariance}
Zhun Zhong, Liang Zheng, Zhiming Luo, Shaozi Li, and Yi Yang.
\newblock Invariance matters: Exemplar memory for domain adaptive person
  re-identification.
\newblock In {\em CVPR}, 2019.

\bibitem{zhou2019iou}
Dingfu Zhou, Jin Fang, Xibin Song, Chenye Guan, Junbo Yin, Yuchao Dai, and
  Ruigang Yang.
\newblock Iou loss for 2{D}/3{D} object detection.
\newblock In {\em 3DV}, 2019.

\bibitem{zhou2018voxelnet}
Yin Zhou and Oncel Tuzel.
\newblock Voxelnet: End-to-end learning for point cloud based 3{D} object
  detection.
\newblock In {\em CVPR}, 2018.

\end{thebibliography}
}
\clearpage


\section*{Supplementary material}

# Supplementary Material
# HVPR: Hybrid Voxel-Point Representation for Single-stage 3D Object Detection

Jongyoun Noh    Sanghoon Lee    Bumsub Ham*

School of Electrical and Electronic Engineering, Yonsei University

## 1. Implementation details

**Point-based and voxel-based features.** For each 3D scene, we subsample 16,384 point clouds, and use them as our inputs. We exploit PointNet++ [4], consisting of set abstraction (SA) and feature propagation (FP) layers, to extract point-based features. Specifically, we use two SA layers to downsample the number of point clouds from 16,384 to 4,096 and 1,024, respectively. We then obtain the point-based features of size $64 \times 1$ for each input point cloud after applying two FP layers. To extract voxel-based features, we use a multi-layer perceptron (MLP) consisting of two hidden layers, whose channel sizes are 32 and 64, respectively.

**Backbone network.** Our backbone network consists of three convolutional blocks, where each block extracts a feature map of a specific scale. See Fig. 2 in the main paper (Backbone Network with AMFM). Each block has three 2D convolutional, BatchNorm [2] and ReLU layers. For the first layer of each block, we exploit a convolution with stride 2 in order to downsample input feature maps. The sizes of output channels for feature maps are 128, 256, and 512, respectively, for each block, and the filter size is set to 3 for all layers.

**AMFM.** We use a MLP with two hidden layers with channel sizes of 16 and 32, respectively, to obtain voxel-wise representations of the 3D scale features. We adopt a $3 \times 3$ convolutional layer to produce a spatial attention map in AMFM. We upsample scale-aware features with a transposed convolution, such that they have the same spatial resolution and channel size as the largest one. See Fig. 2 in the main paper (Backbone Network with AMFM).

**Detection head.** We use two fully connected layers with $384$ channels to localize and classify objects.

**Settings for the pedestrian class.** For the pedestrian class, we train our network for 200 epochs, with a learning rate of 2e-4 and a weight decay of 1e-4. The learning rate is decayed by a factor of $0.8$ every 15 epochs. Batch size is set

*Corresponding author.

Table 1: Runtime analysis for each step of our model.

| Step | Time |
|---|---|
| Data preprocessing | 1.2 ms |
| Voxel encoder | 1.5 ms |
| Pseudo image w/ memory | 3.2 ms |
| Backbone w/ AMFM | 9.9 ms |
| Head and post-processing | 11.9 ms |

to 1 for each GPU. We apply a global translation, where a translation factor is drawn from $\mathcal{N}(0, 0.2)$, in addition to the data augmentation techniques as for the car class. The parameters of our model for the pedestrian class are the same with the ones for the car class except for the number of prototypes ($K = 20$ for car, $K = 10$ for pedestrian). For other parameters, we use the same setting as in [3] as follows: Assuming that the dimension of a 3D scene $(W, H, L)$ is within a range of $[(0, 48), (-20, 20), (-2.5, 0.5)]$ meters. We set the size of anchors to $0.6 \times 0.8 \times 1.73$. We choose the anchors with the IoU scores larger than 0.5 as positive boxes, while those lower than 0.35 are used as negative ones. We use $(0.16, 0.16, 4)$ as the size of a voxel, $v_W \times v_H \times v_L$. The number of point clouds within each voxel $N_{\text{vox}}$ and the size of augmented point clouds $D$ are set to 100 and 9, respectively.

## 2. More results

**Runtime analysis.** The average runtime of our full model for the car class is 27.7 milliseconds with an Nvidia 2080Ti GPU. The detailed runtime for each step is shown in Table 1. We can see that most computation time is spent for the backbone network and the post processing. Estimating a pseudo image just takes 4.7 milliseconds.

**AMFM.** We visualize in Fig. 1 activations of multi-scale and scale-aware features from the first and the second blocks in a backbone network. Note that the scale-aware features are obtained by applying AMFM to multi-scale features. We can see that scale-aware features highly activate on decisive regions, compared to multi-scale ones. More specifically, the scale-aware features from the first block,

which are of high-resolution, activate on small or distant objects, while suppressing the features near a LiDAR sensor. The features from the second block, on the other hand, are of low-resolution. They attend more to regions near the sensor typically containing large objects. This suggests that AMFM allows our network to consider complex scale variations for object localization.

**Qualitative results.** We visualize in Fig. 2 detection results on the validation split of KITTI [1]. We can see that our model localizes small and/or occluded objects with sparse point clouds well. This indicates our hybrid 3D representation and scale-aware features are effective to localize hard examples and robust to complex scale variations. We also show failure examples in the last row of Fig. 2. Our model misses the heavily occluded objects, *e.g.*, in the left and middle images, some of which are not visible even in the RGB image. It also does not localize the object captured with little or no point clouds as shown in the right image.

## References

[1] Andreas Geiger, Philip Lenz, and Raquel Urtasun. Are we ready for autonomous driving? the KITTI vision benchmark suite. In *CVPR*, 2012. 2, 4
[2] Sergey Ioffe and Christian Szegedy. Batch normalization: Accelerating deep network training by reducing internal covariate shift. In *ICML*, 2015. 1
[3] Alex H Lang, Sourabh Vora, Holger Caesar, Lubing Zhou, Jiong Yang, and Oscar Beijbom. Pointpillars: Fast encoders for object detection from point clouds. In *CVPR*, 2019. 1
[4] Charles Ruizhongtai Qi, Li Yi, Hao Su, and Leonidas J Guibas. Pointnet++: Deep hierarchical feature learning on point sets in a metric space. In *NeurIPS*, 2017. 1

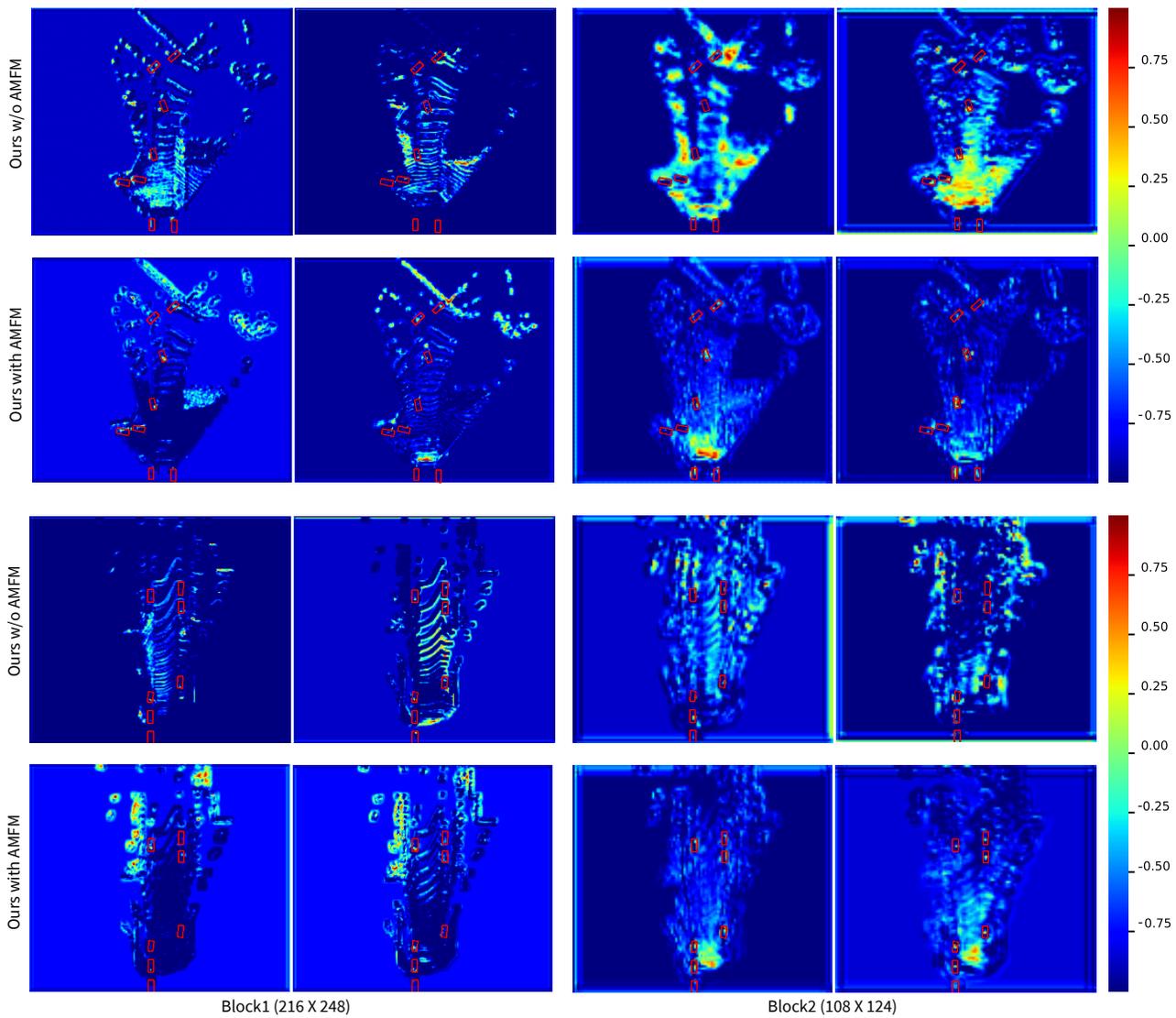

Figure 1: Qualitative comparison of multi-scale and scale-aware features from the first and the second blocks in the backbone network. We visualize feature activations w.r.t input point clouds. The numbers in parentheses indicate a spatial resolution of a feature map, and the red boxes show ground-truth objects. See text for details.

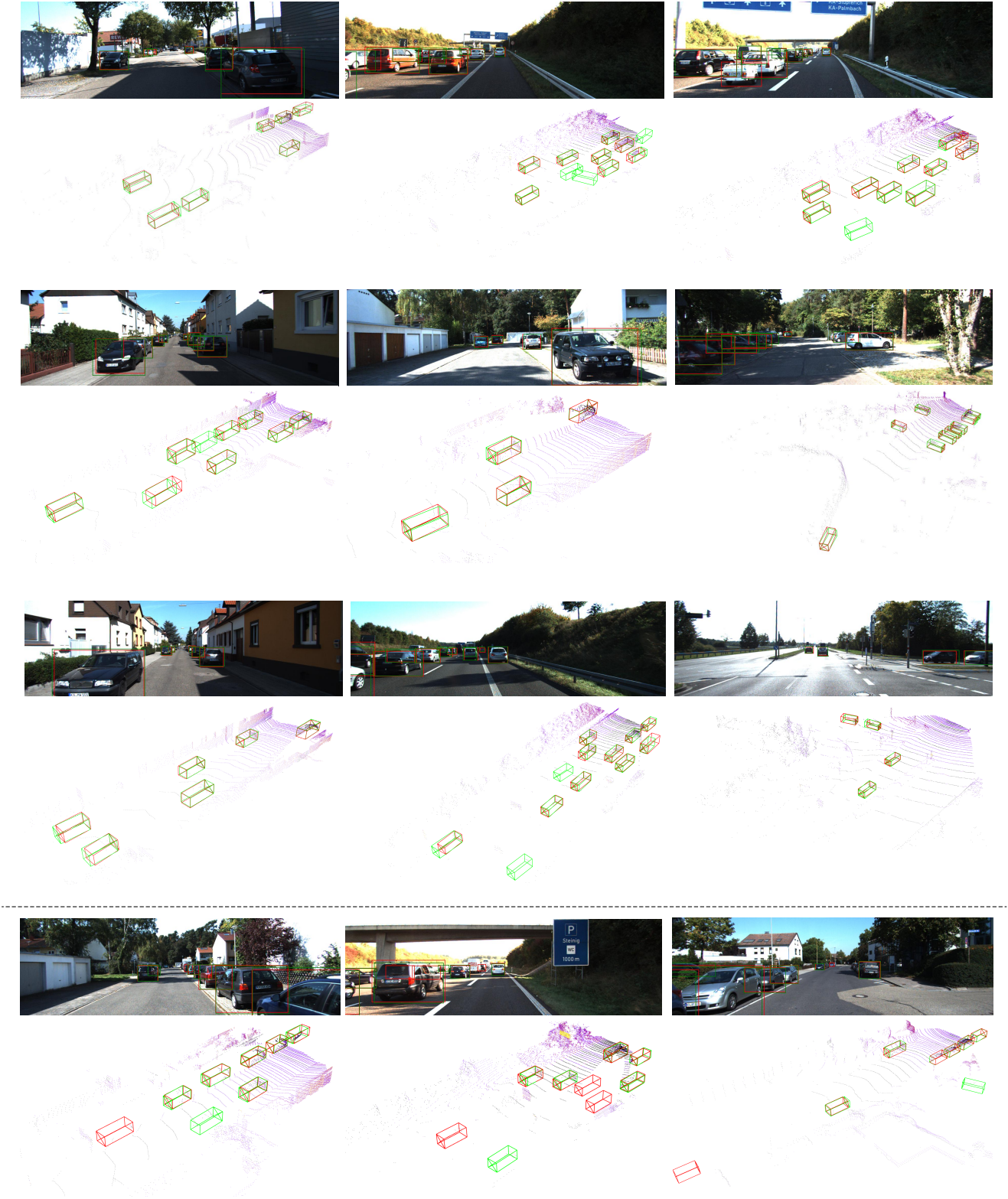

Figure 2: Qualitative results on the validation split of KITTI [1]. Our predictions and ground-truth bounding boxes are shown in green and red, respectively. We also show 2D bounding boxes projected from 3D detection results. The last row shows some failure cases. Best viewed in color.



\end{document}